\documentclass[11pt,a4paper]{article}
\usepackage[hyperref]{acl2020}
\usepackage{times}
\usepackage{latexsym}

\usepackage{url}
\usepackage{multirow}

\usepackage{arydshln}
\usepackage{amsmath}

\usepackage{subfigure}
\usepackage{graphicx}

\usepackage{microtype}

\usepackage{CJKutf8}
\aclfinalcopy 

\newcommand{\Sref}[1]{\S\ref{#1}}

\newcommand{\Fref}[1]{Figure~\ref{#1}}
\newcommand{\tref}[1]{table~\ref{#1}}

\newcommand{\ignore}[1]{}

\title{A Deep Reinforced Model for Zero-Shot Cross-Lingual Summarization \\ with Bilingual Semantic Similarity Rewards}

\author{Zi-Yi Dou ~ Sachin Kumar ~ Yulia Tsvetkov \\
  Language Technologies Institute\\ Carnegie Mellon University \\
  {\tt \{zdou, sachink, ytsvetko\}@cs.cmu.edu}}

\date{}

\begin{document}
\maketitle
\begin{abstract}
Cross-lingual text summarization aims at generating a document summary in one language given input in another language. It is a practically important but under-explored task, primarily due to the dearth of available data. Existing methods resort to machine translation to synthesize training data, but such pipeline approaches suffer from error propagation. In this work, we propose an end-to-end cross-lingual text summarization model. The model uses reinforcement learning to directly optimize a bilingual semantic similarity metric between the summaries generated in a target language and gold summaries in a source language. We also introduce techniques to pre-train the model leveraging monolingual summarization and machine translation objectives. Experimental results in both English--Chinese and English--German cross-lingual summarization settings demonstrate the effectiveness of our methods. In addition, we find that reinforcement learning models with bilingual semantic similarity as rewards generate more fluent sentences than strong baselines.\footnote{\url{https://github.com/zdou0830/crosslingual_summarization_semantic}.}
\end{abstract}

\section{Introduction}
Cross-lingual text summarization (XLS) is the task of compressing
a long article in one language into a summary in a different language. 
Due to the dearth of training corpora, standard sequence-to-sequence approaches to summarization cannot be applied to this task. 
Traditional approaches to XLS thus follow a pipeline, for example, summarizing the article in the source language followed by translating the summary into the target language or vice-versa~\cite{wan2010cross,wan2011using}. Both of these approaches require separately trained summarization and translation models, and suffer from error propagation~\cite{ncls}.

  \begin{figure}[t]%
  \centering
  \includegraphics[width=0.46\textwidth]{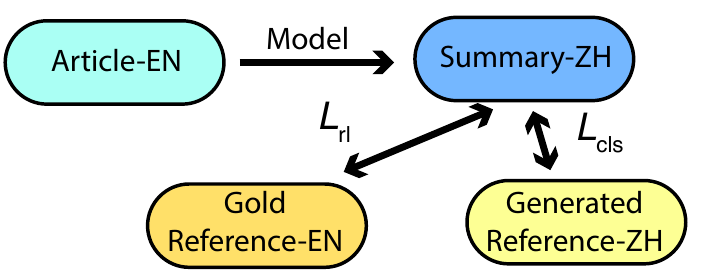}
  \caption{Along with minimizing the XLS cross-entropy loss $L_\mathrm{xls}$, we also apply reinforcement learning to optimize the model by directly comparing the outputs with gold references in the source language.}
  \label{fig:fig1}
  \end{figure}

Prior studies have attempted to train XLS models in an end-to-end fashion, through knowledge distillation from pre-trained machine translation (MT) or monolingual summarization models~\citep{shen2018zero,duan2019zero}, but these approaches have been only shown to work for short outputs. 
Alternatively, \citet{ncls} proposed to automatically translate source-language summaries in the training set thereby generating pseudo-reference summaries in the target language. With this parallel dataset of source documents and target summaries
, an end-to-end model is trained to simultaneously summarize and translate using a multi-task objective. 
Although the XLS model is trained end-to-end, it is trained on MT-generated reference translations and is still prone to compounding of translation and summarization errors. 

In this work, we propose to train an end-to-end XLS model to directly generate target language summaries given the source articles by matching the semantics of the predictions 
with the semantics of the source language summaries. To achieve this, we use reinforcement learning (RL) with a bilingual semantic similarity metric as a reward~\cite{wieting2019simple}. This metric is computed between the machine-generated summary in the target language and the gold summary in the source language. 
Additionally, to better initialize our XLS model for RL, we propose a new multi-task pre-training objective based on machine translation and monolingual summarization to encode common information available from the two tasks. To enable the 
model to still differentiate between the two tasks, we add task specific tags to the input~\cite{wu2016google}.

We evaluate our proposed method on English--Chinese and English--German XLS test sets. These test corpora are constructed by first using an MT-system to translate source summaries to the target language, and then being post-edited by human annotators. 
Experimental results 
demonstrate that just using our proposed pre-training method without fine-tuning with RL improves the best-performing baseline by up to 0.8 ROUGE-L points. Applying reinforcement learning yields further improvements in performance by up to 0.5 ROUGE-L points. Through extensive analyses and human evaluation, 
we show that when the bilingual semantic similarity reward is used, our model generates summaries that are more accurate, longer, more fluent, and more relevant than summaries generated by baselines.

\section{Model}
\label{sec:model}
\begin{figure}[t]
\centering
\includegraphics[width=0.5\textwidth]{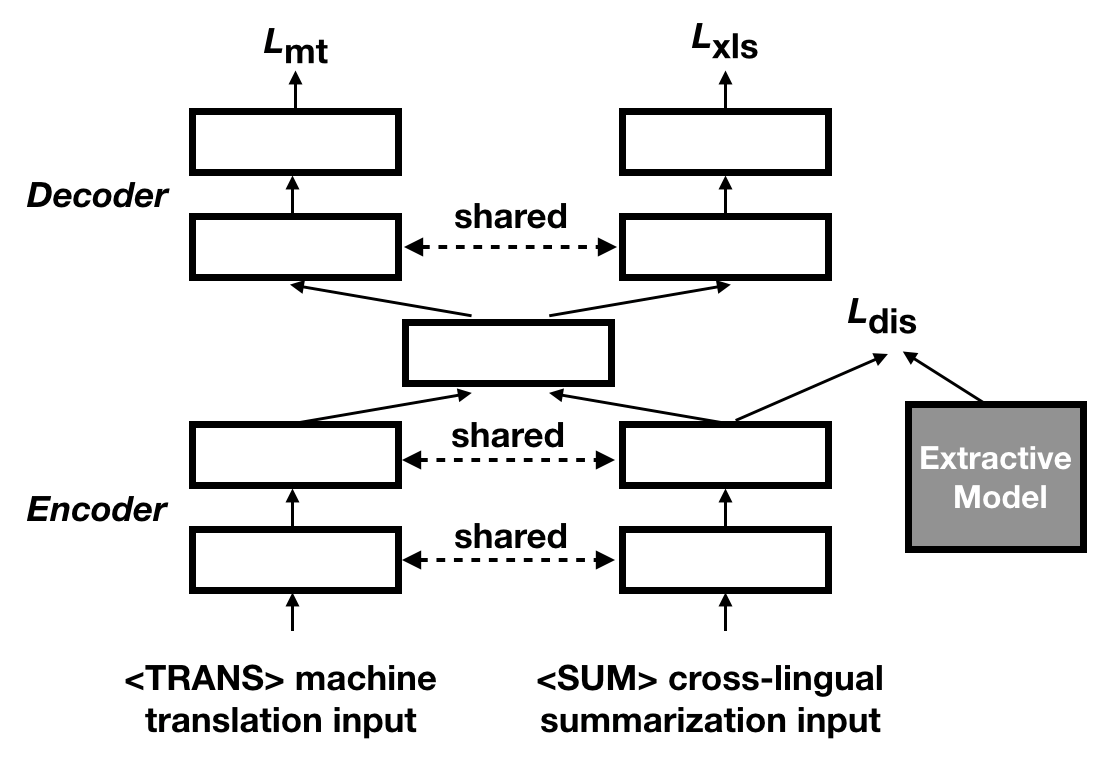}
\caption{Illustration of the supervised pre-training stage. The model is trained with cross-lingual summarization, machine translation and distillation objectives. The parameters of bottom layers of the decoders are shared across tasks.}
\label{fig:arch}
\end{figure}

In this section, we describe the details of the task and our proposed approach. 
\subsection{Problem Description}

We first formalize our task setup. We are given $N$ articles and their summaries in the source language  $\{(x_\mathrm{src}^{(1)}, y_\mathrm{src}^{(1)}), \ldots, (x_\mathrm{src}^{(N)}, y_\mathrm{src}^{(N)})\}$ as a training set. Our goal is to train a summarization model $f(\cdot; \theta)$ which takes an article in the source language $x_\mathrm{src}$ as input and generates its summary in a pre-specified target language $\hat{y}_\mathrm{tgt}=f(x_\mathrm{src}; \theta)$. Here, $\theta$ are the learnable parameters of $f$. During training, no gold summary $y^{(i)}_\mathrm{tgt}$ is available. 

Our model consists of one encoder, denoted as $E$, which takes $x_\mathrm{src}$ as input and generates its vector representation $\mathbf{h}$. $\mathbf{h}$ is fed as input to two decoders. The first decoder $D_1$ predicts the summary in the target language ($\hat{y}_\mathrm{tgt}$) one token at a time. The second decoder $D_2$ predicts the translation of the input text ($\hat{v}_\mathrm{tgt}$). While both $D_1$ and $D_2$ are used during training, only $D_1$ is used for XLS at test time. Intuitively, we want the model to select parts of the input article which might be important for the summary and also translate them into the target language. To bias our model to encode this behavior, we propose the following algorithm for pre-training:
\begin{itemize}
    \item Use a machine translation (MT) model to generate pseudo reference summaries ($\tilde{y}_\mathrm{tgt}$) by translating $y_\mathrm{src}$ to the target language. Then, translate $\tilde{y}_\mathrm{tgt}$ back to the source language using a target-to-source MT model and discard the examples with high reconstruction errors, which are measured with ROUGE~\citep{lin2004rouge} scores. The details of this step can be found in~\citet{ncls}.
    \item Pre-train the model parameters $\theta$ using a multi-task objective based on MT and monolingual summarization objectives with some simple yet effective techniques as described in~\Sref{subsec:pretrain}.
    \item Further fine-tune the model using reinforcement learning with bilingual semantic similarity metric~\citep{wieting2019simple} as reward, which is described in~\Sref{subsec:rl}.
\end{itemize}

\subsection{Supervised Pre-Training Stage}
\label{subsec:pretrain}
Here, we describe the second step of our algorithm (Figure \ref{fig:arch}). The pre-training loss we use is a weighted combination of three objectives. Similarly  to~\citet{ncls}, we use an XLS pre-training objective and an MT pre-training objective as described below with some simple but effective improvements. We also introduce an additional objective based on distilling knowledge from a monolingual summarization model.

\paragraph{XLS Pre-training Objective ($L_{\mathrm{xls}}$)}
This objective computes the cross-entropy loss of the predictions from $D_1$, considering the machine-generated summaries in the target language, $\tilde{y}_\mathrm{tgt}^{(i)}$ as references, given $x_\mathrm{src}^{(i)}$ as inputs. Per sample, this loss can be formally written as:
$$L_{\mathrm{xls}} = \sum_{j=1}^M \log p(\tilde{y}_{\mathrm{tgt},j}^{(i)} | \tilde{y}_{\mathrm{tgt},<j}^{(i)}, x_\mathrm{src}^{(i)})$$ where M is the number of tokens in the summary $i$. 

\paragraph{Joint Training with Machine Translation}
\citet{ncls} argue that machine translation can be considered a special case of XLS with a compression ratio of 1:1. 
In line with~\citet{ncls}, we train $E$ and $D_2$ as the encoder and decoder of a translation model using an MT parallel corpus $\{(u_\mathrm{src}^{(i)}, v_\mathrm{tgt}^{(i)} ) \}$. The goal of this step is to make the encoder have an inductive bias towards encoding information specific to translation. Similar to $L_{\mathrm{xls}}$, the machine translation objective per training sample $L_{\mathrm{mt}}$ is:
$$L_{\mathrm{mt}} = \sum_{j=1}^K \log p (v_{\mathrm{tgt},j}^{(i)} | v_{\mathrm{tgt},<j}^{(i)}, u_\mathrm{src}^{(i)})$$ where $K$ is the number of tokens in $v_\mathrm{tgt}^{(i)}$. 
The $L_{\mathrm{xls}}$ and $L_{\mathrm{mt}}$ objectives are inspired by~\citet{ncls}. We propose the following two enhancements to the model to leverage better the two objectives: 
\begin{enumerate}
    \item We share the parameters of bottom layers of the two decoders, namely $D_1$ and $D_2$, to share common high-level representations while the parameters of the top layers more specialized to decoding are separately trained.
    \item We append an artificial task tag $\langle$\textsc{sum}$\rangle$ (during XLS training) and $\langle$\textsc{trans}$\rangle$ (during MT training) at the beginning of the input document to make the model aware of which kind of input it is dealing with. 
\end{enumerate}
We show in \Sref{subsec:analysis} that such simple modifications result in noticeable performance improvements.


\paragraph{Knowledge Distillation from Monolingual Summarization}
\label{know-distill}

To bias the encoder to identify sentences which can be most relevant to the summary, first, we use an extractive monolingual summarization method to predict the probability $q_i$ of each sentence or keyword in the input article being relevant to the summary. We then distill knowledge from this model into the encoder $E$ by making it predict these probabilities. 

Concretely, we append an additional output layer to the encoder of our model and it predicts the probability $p_i$ of including the $i$-th sentence or word in the summary. The objective is to minimize the difference between $p_i$ and $q_i$. We use the following loss (for each sample) for the model encoder:\footnote{We also experimented with a common distillation objective based on minimizing KL divergence, $\frac{1}{n}\sum_{i=1}^n q_i \log p_i$, but it did not perform as well.}

\begin{equation}
\label{eqn:dis}
L_{\mathrm{dis}} = \frac{1}{L}\sum_{j=1}^L (\log q_j - \log p_j)^2,
\end{equation}
where $L$ is the number of sentences or keywords in each article. 

Our final pre-training objective during the supervised pre-training stage is:
\begin{equation}
    L_{\mathrm{sup}} = L_{\mathrm{xls}} + L_{\mathrm{mt}} + \lambda L_{\mathrm{dis}}
\label{eq:pretrain}
\end{equation}
where
$\lambda$ is a hyper-parameter and is set to 10 in our experiments. Training with $L_\mathrm{mt}$ requires an MT parallel corpus whereas the other two objectives utilize the cross-lingual summarization dataset. Pre-training algorithm alternates between the two parts of the objective using mini-batches from the two datasets as follows until convergence:

\begin{enumerate}
    \item Sample a minibatch from the MT corpus $\{(u_\mathrm{src}^{(i)}, v_\mathrm{tgt}^{(i)})\}$ and train the parameters of $E$ and $D_2$ with $L_\mathrm{mt}$.
    \item Sample a minibatch from the XLS corpus, $\{(x_\mathrm{src}^{(i)}, \tilde{y}_\mathrm{tgt}^{(i)})\}$ and train the parameters of $E$ and $D_1$ with $L_\mathrm{xls} + \lambda L_\mathrm{dis}$.
\end{enumerate}

\subsection{Reinforcement Learning Stage}
\label{subsec:rl}

For XLS, the target language reference summaries ($\tilde{y}_\mathrm{tgt}$) used during pre-training are automatically generated with MT models and thus they may contain errors. In this section, we describe how we further fine-tune the model using only human-generated source language summaries ($y_\mathrm{src}$) with reinforcement learning (RL). Specifically, we first feed the article $x_\mathrm{src}$ as an input to the encoder $E$, and generate the target language summary $\hat{y}_\mathrm{tgt}$ using $D_1$. We then compute a cross-lingual similarity metric between $\hat{y}_\mathrm{tgt}$ and $y_\mathrm{src}$ and use it as a reward to fine-tune $E$ and $D_1$.

Following~\citet{paulus2017deep}, we adopt two different strategies to generate $\hat{y}_\mathrm{tgt}$ at each training iteration, (a) $\hat{y}_\mathrm{tgt}^{s}$ obtained by sampling from the softmax layer at each decoding step, and (b) $\hat{y}_\mathrm{tgt}^{g}$ obtained by greedy decoding. The RL objective per sample is given by:
\begin{equation}
\label{eqn:rl}
L_{\mathrm{rl}} = \left( r(\hat{y}_\mathrm{tgt}^{g})-r(\hat{y}_\mathrm{tgt}^{s}) \right) \sum_{j=1}^M \log p (\hat{y}_\mathrm{tgt,i}^{s} | \hat{y}_\mathrm{tgt,<j}^{s}, x),
\end{equation}
where $r(\cdot)$ is the reward function. To fine-tune the model, we use the following hybrid training objective: $\gamma L_{\mathrm{rl}} + (1-\gamma) L_{\mathrm{xls}}$, where
$\gamma$ is a scaling factor.

We train a cross-lingual similarity model (\textsc{xsim}) with the best performing model in~\citet{wieting2019simple}. This model is trained using an MT parallel corpus. Using \textsc{xsim}, we obtain sentence representations for both $\hat{y}_\mathrm{tgt}$ and $y_\mathrm{src}$ and treat the cosine similarity between the two representations as the reward $r(\cdot)$. 

\begin{table*}[t]
  \centering
   \resizebox{1.0\textwidth}{!}{%
  \begin{tabular}{l|llll|llll}
   \multicolumn{1}{c|}{\multirow{2}{*}{\bf Method} } &\multicolumn{4}{c|}{\bf English--Chinese} & \multicolumn{4}{c}{\bf English--German} \\
  \cline{2-9}
 & \textbf{ROUGE-1} & \textbf{ROUGE-2} & \textbf{ROUGE-L} & 
  \textbf{\textsc{xsim}} & \textbf{ROUGE-1} & \textbf{ROUGE-2} & \textbf{ROUGE-L} & 
  \textbf{\textsc{xsim}} \\
   \hline
   \multicolumn{4}{l}{\it Pipeline-Based Methods} \\
   \hline
   \textsc{Tran-Sum}~\citep{ncls} & 28.19 & 11.40 & 25.77 & -& - & - & - & -\ \\
   \textsc{Sum-Tran}~\citep{ncls} & 32.17 & 13.85 & 29.43 & - & - & - & - & -\\
      \hline
   \multicolumn{4}{l}{\it End-to-End Training Methods} \\
   \hline
   \textsc{MLE-xls} & 37.38 & 17.96 & 33.85 & 45.17 & 23.06 & 8.40 & 21.28 & 43.41\\
   \textsc{MLE-xls+mt}~\citep{ncls} & 40.23 & 22.32 & 36.59 & - & - & - & - & -\\
   \textsc{MLE-xls+mt} (Reimplemented) & 41.25 & 22.40 & 37.93 & 48.77 & 36.83 & 17.62 & 35.54 & 49.21 \\
    \hline
   \textsc{MLE-xls+mt+dis} & 42.19* & 22.91* & 38.74* & 49.20* & 37.74* & 18.40* & 36.34* & 49.53*\\
   \hdashline
   \textsc{RL-rouge} & 42.51* & 22.96 & 38.98* & 49.65 & 38.32* & 18.46 & 36.86* & 49.66\\
   \textsc{RL-xsim} & \bf 42.83* & \bf 23.30* & \bf 39.29* & \bf 50.85* & \bf 38.69* & \bf 18.76* & \bf 37.20* & \bf 50.17* \\
   \textsc{RL-rouge+xsim} & 42.49 & 23.29* & 38.95 & 49.88*  & 38.19* & 18.17 & 36.72* & 49.64 \\
  \hline
    \end{tabular}
    }
    \caption{ \label{tab:main1} Performance of different models. The highest scores are in {\bf bold} and statistical significance compared with the best baseline is indicated with * ($p<$0.05, computed using {\it compare-mt}~\cite{neubig2019compare}). \textsc{Xsim} is computed between the target language system outputs and the source language reference summaries.}
  \end{table*}
  
\section{Experimental Setup}

\subsection{Datasets} We evaluate our models on English--Chinese and English--German article-summary datasets. The English--Chinese dataset is created by~\citet{ncls}, constructed using the CNN/DailyMail monolingual summarization corpus~\cite{hermann2015teaching}. The training, validation and test sets consist of about 364K, 3K and 3K samples, respectively. The English--German dataset is our contribution, constructed from the Gigaword dataset~\cite{rush2015neural}. We sample 2.48M training, 2K validation and 2K test samples from the dataset. Pseudo-parallel corpora for both language pairs are constructed by translating the summaries to the target language (and filtered after back-translation; see \Sref{sec:model}). This is done for training, validation as well as test sets.  
These two pseudo-parallel training sets are used for pre-training with $L_\mathrm{xls}$.
Translated Chinese and German summaries of the test articles are then post-edited by human annotators to construct the test set for evaluating XLS.
We refer the readers to~\citep{ncls} for more details. 
For the English--Chinese dataset, we use word-based segmentation for the source (articles in English) and character-based segmentation for the target (summaries in Chinese) as in~\citep{ncls}. For the English--German dataset, byte-pair encoding is used~\cite{sennrich2016neural} with 60K merge operations. For machine translation and training the \textsc{xsim} model, we sub-sample 5M sentences from the WMT2017 Chinese--English and WMT2014 German--English training dataset~\cite{bojar2014findings,bojar2017findings}. 

\subsection{Implementation Details} 
\label{subsec:implement_details}
We use the Transformer-\textsc{base} model~\citep{vaswani2017attention} as the underlying architecture for our model ($E$, $D_1$, $D_2$, extractive summarization model for distillation and baselines). We refer the reader to~\citet{vaswani2017attention} for hyperparameter details. In the input article, a special token $\langle$\textsc{SEP}$\rangle$ is added at the beginning of each sentence to mark sentence boundaries. 
For the CNN/DailyMail corpus, the monolingual extractive summarization used in the distillation objective has the same architecture as the encoder $E$ and is trained the CNN/DailyMail corpus constructed by~\citep{liu2019text}. To train the encoder with $L_\mathrm{dis}$, we take the final hidden representation of each $\langle$\textsc{SEP}$\rangle$ token and apply a 2-layer feed-forward network with ReLU activation in the middle layer and sigmoid at the final layer to get $q_i$ for each sentence $i$ (see \Sref{know-distill}).

For the Gigaword dataset, because the inputs and outputs are typically short, we choose keywords rather than sentences as the prediction unit. Specifically, we first use TextRank~\cite{mihalcea2004textrank} to extract all the keywords from the source document. Then, for each keyword $i$ that appears in the target summary, the gold label $q_i$ in equation~\ref{eqn:dis} is assigned to 1, and $q_i$ is assigned to 0 for keywords that do not appear in the target side. 

We share the parameters of the bottom four layers of the decoder in the multi-task setting. We use the \textsc{Trigram} model in~\citep{wieting2019simple, wieting2019beyond} to measure the cross-lingual sentence semantic similarities. As pointed out in \Sref{sec:model}, after the pre-training stage, we only use $D_1$ for XLS. The final results are obtained using only $E$ and $D_1$. We use two metrics for evaluating the performance of models: ROUGE (1, 2 and L)~\cite{lin2004rouge} and \textsc{xsim}~\cite{wieting2019simple}. 

 Following~\newcite{paulus2017deep}, we select $\gamma$ in equation \ref{eqn:rl} to
$0.998$ for the Gigaword Corpus and $\gamma= 0.9984$ for the
CNN/DailyMail dataset.

\subsection{Baselines} We compare our proposed method with the following baselines:

\paragraph{Pipeline Approaches} We report results of summarize-then-translate (\textsc{Sum-Tran}) and translate-then-summarize (\textsc{Tran-Sum}) pipelines. These results are taken from~\citet{ncls}. 
\paragraph{\textsc{MLE-xls}} We pre-train $E$ and $D_1$ with only $L_\mathrm{xls}$ without any fine-tuning.
\paragraph{\textsc{MLE-xls+mt}} We pre-train $E$, $D_1$ and $D_2$ with $L_\mathrm{xls}+L_\mathrm{mt}$ without using $L_\mathrm{dis}$. This is the best performing model in~\citep{ncls}. We show their reported results as well as results from our re-implementation.
\paragraph{\textsc{MLE-xls+mt+dis}} We pre-train the model using \eqref{eq:pretrain} without fine-tuning with RL. We also share the decoder layers and add task specific tags 
to the input as described in \Sref{subsec:pretrain}. 
\paragraph{\textsc{RL-rouge}} Using ROUGE score as a reward function has been shown to improve summarization quality for monolingual summarization models~\citep{paulus2017deep}. In this baseline, we fine-tune the pre-trained model in the above baseline using ROUGE-L as a reward instead of the proposed \textsc{xsim}. The ROUGE-L score is computed between the output of $D_1$ and the machine-generated summary $\tilde{y}_\mathrm{tgt}$. 

\paragraph{\textsc{RL-rouge+xsim}} Here, we use the average of ROUGE score and \textsc{xsim} score as a reward function to fine-tune the pre-trained model (\textsc{MLE-xls+mt+dis}).

\begin{table}[t]
  \centering
   \resizebox{0.45\textwidth}{!}{%
  \begin{tabular}{llll}
  \multicolumn{1}{c}{ \bf Method} & \bf ROUGE-1 & \bf ROUGE-2 & \bf ROUGE-L \\
   \hline
   \textsc{MLE-xls} & 37.38 & 17.96 & 33.85 \\
   +\textsc{Extract} & 39.19 & 19.58 & 35.68 \\
   +\textsc{Dis} & {\bf 40.46} & {\bf 20.47} & {\bf 36.93}  \\
   \hline
   \textsc{MLE-xls+mt} & 41.25 & 22.40 & 37.93 \\
   +\textsc{Extract} & 40.04 & 21.58 & 36.74 \\
   +\textsc{Dis} & \bf 41.68 & \bf 22.48 & \bf 38.29 \\
  \hline
    \end{tabular}
    }
    \caption{\label{tab:extract}  Effect of using hard (\textsc{Extract}) vs soft (\textsc{Dis}) extraction  of summary sentences from the input article}
  \end{table}
  
\section{Results}
\label{section:results}
  The main results of our experiments are summarized in Table~\ref{tab:main1}. Pipeline approaches, as expected, show the weakest performance, lagging behind even the weakest end-to-end approach by more than 5 ROUGE-L points. \textsc{Tran-Sum} performs even worse than \textsc{Sum-Tran}, likely because the translation model is trained on sentences and not long articles. First translating the article with many sentences introduces way more errors than translating a short summary with fewer sentences would. Using just our pre-training method as described in \ref{subsec:pretrain} (\textsc{MLE-xls+mt+dis}), our proposed model out-performs the strongest baseline (\textsc{MLE-xls+mt}) in both ROUGE-L (by 0.8) and \textsc{xsim} (by 0.5). 
  Applying reinforcement learning to fine-tune the model with both ROUGE (\textsc{RL-rouge}), \textsc{xsim} (\textsc{RL-xsim}) or their mean (\textsc{RL-rouge+xsim}) as rewards results in further improvements. Our proposed method, \textsc{RL-xsim} performs the best overall, indicating the importance of using cross-lingual similarity as a reward function. \textsc{RL-rouge} uses a machine-generated reference to compute the rewards since target language summaries are unavailable, which might be a reason for its worse performance.
  

\begin{figure*}[t]
\centering
\includegraphics[width=0.95\textwidth]{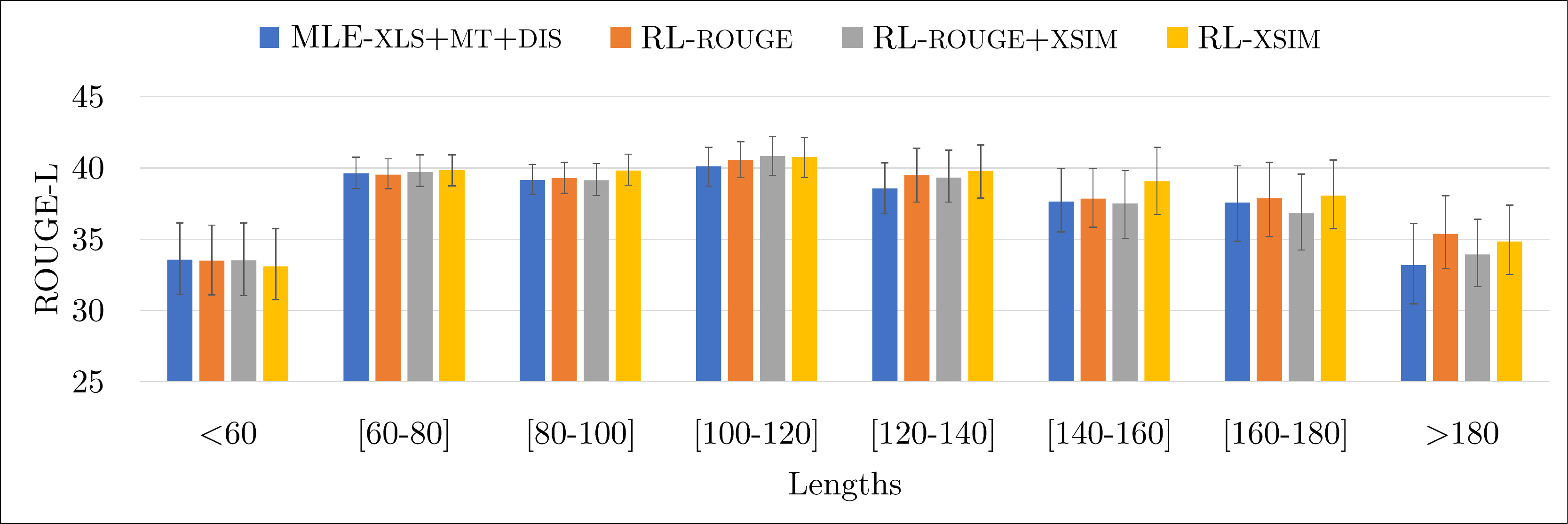}
\caption{Reinforcement learning can make the model better at generating long summaries. We use the {\it compare-mt} tool~\cite{neubig2019compare} to get these statistics.}
\label{fig:len}
\end{figure*}
\subsection{Analysis}
\label{subsec:analysis}

In this section, we conduct experiments on the CNN/DailyMail dataset to establish the importance of every part of the proposed method and gain further insights into our model.

\paragraph{Soft Distillation vs. Hard Extraction} The results in ~\tref{tab:main1} already show that adding the knowledge distillation objective $L_\mathrm{dis}$ to the pre-training leads to an improvement in performance. The intuition behind using $L_\mathrm{dis}$ is to bias the model to (softly) select sentences in the input article that might be important for the summary. Here, we replace this soft selection with a hard selection. That is, using the monolingual extractive summarization model (as described in \Sref{subsec:implement_details}), we extract top 5 sentences from the input article and use them as the input to the encoder instead. We compare this method with $L_\mathrm{dis}$ as shown in Table~\ref{tab:extract}. With just MLE-\textsc{XLS} as the pre-training objective, \textsc{Extract} shows improvement (albeit with lower overall numbers) in performance but leads to a decrease in performance of MLE-$\textsc{XLS+MT}$. On the other hand, using the distillation objective helps in both cases. 

\paragraph{Effect of the Sharing and Tagging Techniques} In Table~\ref{tab:sharetag}, we demonstrate that introducing simple enhancements like sharing the lower-layers of the decoder (share) and adding task-specific tags (tags) during multi-task pre-training also helps in improving the performance while at the same using fewer parameters and hence a smaller memory footprint. 

\begin{table}[t]
  \centering
   \resizebox{0.45\textwidth}{!}{%
  \begin{tabular}{llll}
  \multicolumn{1}{c}{ \bf Method} & \textbf{ROUGE-1} & \textbf{ROUGE-2} & \textbf{ROUGE-L} \\
   \hline
   \textsc{MLE+xls+mt} & 41.25 & 22.40 & 37.93 \\
   +\textsc{Share} & 41.36 & 22.43 & 37.95 \\
   +\textsc{Tag} & 41.45 & 22.47 & 38.04 \\
  \hline
    \end{tabular}
    }
    \caption{ \label{tab:sharetag} Effect of sharing decoder layers and adding task-specific tags}
  \end{table}

\paragraph{Effect of Summary Lengths} Next, we study how different baselines and our model performs with respect to generating summaries (in Chinese) of different lengths, in terms of number of characters.
As shown in Figure~\ref{fig:len}, after fine-tuning the model with RL, our proposed model becomes better at generating longer summaries than the one with only pre-training (referred to as \textsc{MLE-xls+mt+dis} in the figure) with \textsc{RL-xsim} performing the best in most cases. We posit that this improvement is due to RL based fine-tuning reducing the problem of exposure bias introduced during teacher-forced pre-training, which especially helps longer generations. 
  
\begin{figure*}[t]
\centering
\includegraphics[width=1.0\textwidth]{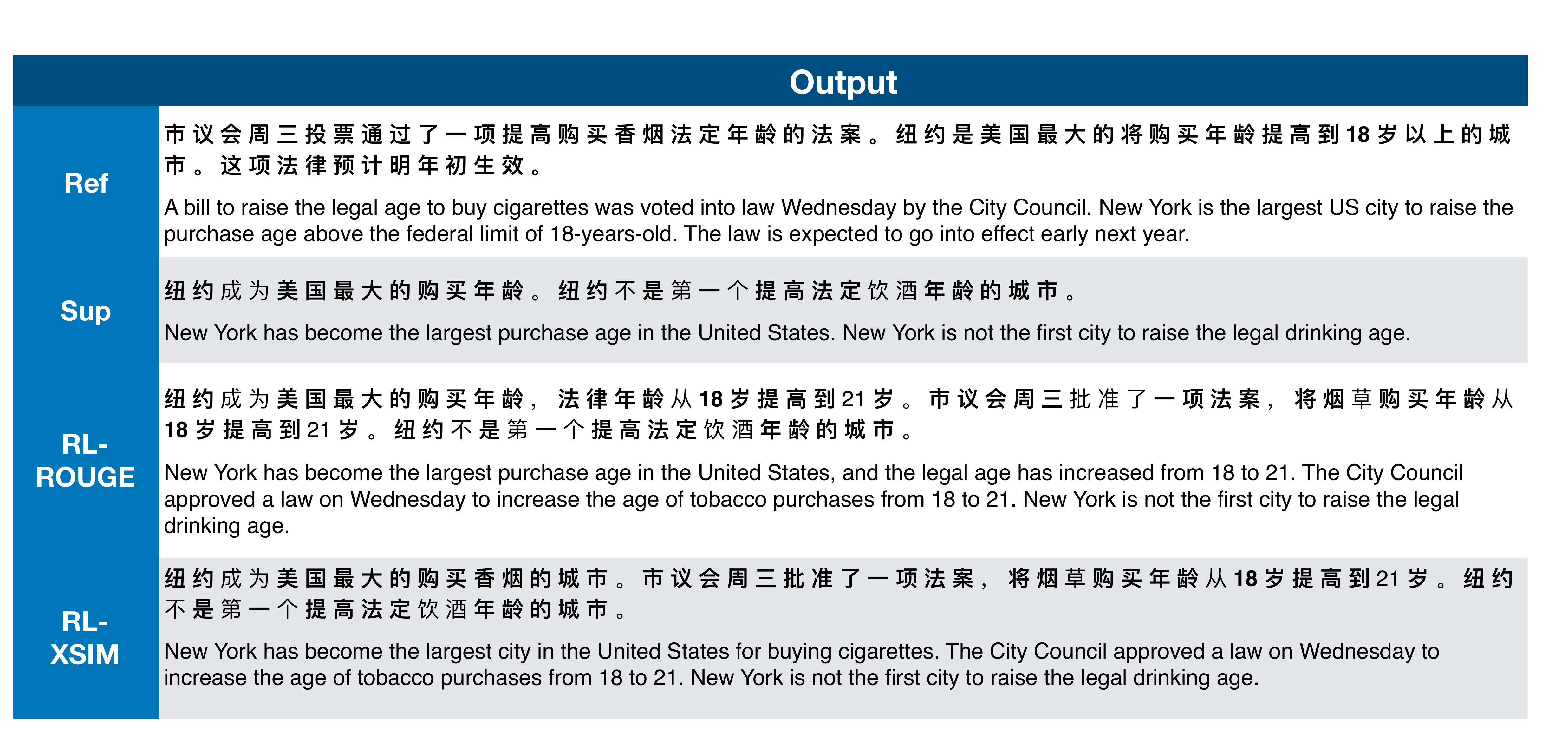}
\caption{Example outputs. The bilingual semantic similarity rewards can make the output more fluent than using ROUGE-L as rewards. ``Sup'' refers to the \textsc{MLE-xls+mt+dis} baseline.}
\label{fig:example}
\end{figure*}

\paragraph{Human Evaluation} In addition to automatic evaluation, which can sometimes be misleading, we perform human evaluation of summaries generated by our models.
 We randomly sample 50 pairs of the model outputs from the test set and ask three human evaluators to compare the pre-trained supervised learning model and reinforcement learning models in terms of \textit{relevance} and \textit{fluency}. For each pair, the evaluators are asked to pick one out of: first model (\textsc{MLE-xls+mt+dis}; lose) , second model(RL models; win) or say that they prefer both or neither (tie). The results are summarized in~\tref{tab:human}. We observe that the outputs of model trained with ROUGE-L rewards are more favored than the ones generated by only pre-trained model in terms of relevance but not fluency. This is likely because the \textsc{RL-rouge} model is trained using machine-generated summaries as references which might lack fluency. \Fref{fig:example} displays one such example. 
On the other hand, cross-lingual semantic similarity as a reward results in generations which are more favored both in terms of relevance and fluency.

 \begin{table}[t]
  \centering
   \resizebox{0.45\textwidth}{!}{%
  \begin{tabular}{lllll}
   \bf Metric & \textbf{Model v. \textsc{MLE}} & \textbf{Win (\%)} & \textbf{Lose (\%)} & \textbf{Tie (\%)}\\
   \hline
   \multirow{2}{*}{Relevance} & \textsc{RL-rouge} & 25.3 &  15.3 & 59.3 \\
   & \textsc{RL-xsim} & 36.0 & 31.3 & 32.7 \\
   \hline
   \multirow{2}{*}{Fluency} & \textsc{RL-rouge} & 13.3 & 17.3 & 69.3 \\
   & \textsc{RL-xsim} & 37.3  & 28.7 & 34.0 \\
   \hline
    \end{tabular}
    }
    \caption{ \label{tab:human} Results showing preferences of human evaluators towards the summaries generated by the mentioned RL methods vs ones from the pre-trained model (\textsc{MLE-xls+mt+dis} referred in short as \textsc{MLE})}
  \end{table}

\section{Related Work}

Most previous work on cross-lingual text summarization utilize either the summarize-then-translate or translate-then-summarize pipeline~\cite{wan2010cross,wan2011using,yao2015phrase,ouyang2019robust}. These methods suffer from error propagation and we have demonstrated their sub-optimal performance in our experiments. Recently, there has been some work on training models for this task in an end-to-end fashion~\citep{shen2018zero,duan2019zero,ncls}, but these models are trained with cross-entropy using machine-generated summaries as references which have already lost some information in the translation step.

Prior work in monolingual summarization have explored hybrid extractive and abstractive summarization objectives which inspires our distillation objective ~\citep{gehrmann2018bottom,hsu2018unified,chen2018fast}. This line of research mainly focus on either compressing sentences extracted by a pre-trained model or biasing the prediction towards certain words.

Language generation models trained with cross-entropy using teacher-forcing suffer from exposure bias and a mismatch between training and evaluation objective. To solve these issues, using reinforcement learning to fine-tune such models have been explored for monolingual summarization where ROUGE rewards is typically used~\citep{paulus2017deep,liu2018generative,pasunuru2018multi}. Other rewards such as BERT score~\citep{zhang2019bertscore} have also been explored~\citep{li2019deep}. Computing such rewards requires access to the gold summaries which are typically unavailable for cross-lingual summarization. This work is the first to explore using cross-lingual similarity as a reward to work around this issue.

\section{Conclusion}
In this work, we propose to use reinforcement learning with a bilingual semantic similarity metric as rewards for cross-lingual document summarization. We demonstrate the effectiveness of the proposed approach in a resource-deficient setting, where target language gold summaries are not available. We also propose simple strategies to better initialize the model towards reinforcement learning by leveraging machine translation and monolingual summarization. 
In future work, we plan to explore methods for stabilizing reinforcement learning as well to extend our methods to other datasets and tasks, such as using the bilingual similarity metric as a reward to improve the quality of machine translation.

\section*{Acknowledgements}
We are grateful to Junnan Zhu, John Wieting, Nikolai Vogler, Graham Neubig for their helpful suggestions and Chunting Zhou, Shuyan Zhou for proofreading the paper. We also thank Ruihan Zhai, Zhi-Hao Zhou for the help with human evaluation and Anurag Katakkar for post-editing the German-English dataset.
This material is based upon work supported by NSF grants IIS1812327 and by Amazon MLRA award. We also thank Amazon for providing GPU credits.

\bibliography{acl2020}

\begin{thebibliography}{28}
\expandafter\ifx\csname natexlab\endcsname\relax\def\natexlab#1{#1}\fi

\bibitem[{Ayana et~al.(2018)Ayana, Shen, Chen, Yang, Liu, and
  Sun}]{shen2018zero}
Ayana, Shi-qi Shen, Yun Chen, Cheng Yang, Zhi-yuan Liu, and Mao-song Sun. 2018.
\newblock Zero-shot cross-lingual neural headline generation.
\newblock \emph{IEEE/ACM Transactions on Audio, Speech and Language
  Processing}.

\bibitem[{Bojar et~al.(2014)Bojar, Buck, Federmann, Haddow, Koehn, Leveling,
  Monz, Pecina, Post, Saint-Amand et~al.}]{bojar2014findings}
Ond{\v{r}}ej Bojar, Christian Buck, Christian Federmann, Barry Haddow, Philipp
  Koehn, Johannes Leveling, Christof Monz, Pavel Pecina, Matt Post, Herve
  Saint-Amand, et~al. 2014.
\newblock Findings of the 2014 workshop on statistical machine translation.
\newblock In \emph{Proc. WMT}.

\bibitem[{Bojar et~al.(2017)Bojar, Chatterjee, Federmann, Graham, Haddow,
  Huang, Huck, Koehn, Liu, Logacheva et~al.}]{bojar2017findings}
Ond{\v{r}}ej Bojar, Rajen Chatterjee, Christian Federmann, Yvette Graham, Barry
  Haddow, Shujian Huang, Matthias Huck, Philipp Koehn, Qun Liu, Varvara
  Logacheva, et~al. 2017.
\newblock Findings of the 2017 conference on machine translation.
\newblock In \emph{Proc. WMT}.

\bibitem[{Chen and Bansal(2018)}]{chen2018fast}
Yen-Chun Chen and Mohit Bansal. 2018.
\newblock Fast abstractive summarization with reinforce-selected sentence
  rewriting.
\newblock In \emph{Proc. ACL}.

\bibitem[{Duan et~al.(2019)Duan, Yin, Zhang, Chen, and Luo}]{duan2019zero}
Xiangyu Duan, Mingming Yin, Min Zhang, Boxing Chen, and Weihua Luo. 2019.
\newblock Zero-shot cross-lingual abstractive sentence summarization through
  teaching generation and attention.
\newblock In \emph{Proc. ACL}.

\bibitem[{Gehrmann et~al.(2018)Gehrmann, Deng, and Rush}]{gehrmann2018bottom}
Sebastian Gehrmann, Yuntian Deng, and Alexander Rush. 2018.
\newblock Bottom-up abstractive summarization.
\newblock In \emph{Proc. EMNLP}.

\bibitem[{Hermann et~al.(2015)Hermann, Kocisky, Grefenstette, Espeholt, Kay,
  Suleyman, and Blunsom}]{hermann2015teaching}
Karl~Moritz Hermann, Tomas Kocisky, Edward Grefenstette, Lasse Espeholt, Will
  Kay, Mustafa Suleyman, and Phil Blunsom. 2015.
\newblock Teaching machines to read and comprehend.
\newblock In \emph{Proc. NeurIPS}.

\bibitem[{Hsu et~al.(2018)Hsu, Lin, Lee, Min, Tang, and Sun}]{hsu2018unified}
Wan-Ting Hsu, Chieh-Kai Lin, Ming-Ying Lee, Kerui Min, Jing Tang, and Min Sun.
  2018.
\newblock A unified model for extractive and abstractive summarization using
  inconsistency loss.
\newblock In \emph{Proc. ACL}.

\bibitem[{Li et~al.(2019)Li, Lei, Qin, and Wang}]{li2019deep}
Siyao Li, Deren Lei, Pengda Qin, and William~Yang Wang. 2019.
\newblock Deep reinforcement learning with distributional semantic rewards for
  abstractive summarization.
\newblock In \emph{Proc. EMNLP}.

\bibitem[{Lin(2004)}]{lin2004rouge}
Chin-Yew Lin. 2004.
\newblock {ROUGE}: A package for automatic evaluation of summaries.
\newblock In \emph{Text Summarization Branches Out}.

\bibitem[{Liu et~al.(2018)Liu, Lu, Yang, Qu, Zhu, and Li}]{liu2018generative}
Linqing Liu, Yao Lu, Min Yang, Qiang Qu, Jia Zhu, and Hongyan Li. 2018.
\newblock Generative adversarial network for abstractive text summarization.
\newblock In \emph{Proc. AAAI}.

\bibitem[{Liu and Lapata(2019)}]{liu2019text}
Yang Liu and Mirella Lapata. 2019.
\newblock Text summarization with pretrained encoders.
\newblock In \emph{Proc. EMNLP}.

\bibitem[{Mihalcea and Tarau(2004)}]{mihalcea2004textrank}
Rada Mihalcea and Paul Tarau. 2004.
\newblock Textrank: Bringing order into text.
\newblock In \emph{Proc. EMNLP}.

\bibitem[{Neubig et~al.(2019)Neubig, Dou, Hu, Michel, Pruthi, and
  Wang}]{neubig2019compare}
Graham Neubig, Zi-Yi Dou, Junjie Hu, Paul Michel, Danish Pruthi, and Xinyi
  Wang. 2019.
\newblock compare-mt: A tool for holistic comparison of language generation
  systems.
\newblock In \emph{Proc. NAACL Demo}.

\bibitem[{Ouyang et~al.(2019)Ouyang, Song, and McKeown}]{ouyang2019robust}
Jessica Ouyang, Boya Song, and Kathleen McKeown. 2019.
\newblock A robust abstractive system for cross-lingual summarization.
\newblock In \emph{Proc. NAACL}.

\bibitem[{Pasunuru and Bansal(2018)}]{pasunuru2018multi}
Ramakanth Pasunuru and Mohit Bansal. 2018.
\newblock Multi-reward reinforced summarization with saliency and entailment.
\newblock In \emph{Proc. NAACL}.

\bibitem[{Paulus et~al.(2018)Paulus, Xiong, and Socher}]{paulus2017deep}
Romain Paulus, Caiming Xiong, and Richard Socher. 2018.
\newblock A deep reinforced model for abstractive summarization.
\newblock In \emph{Proc. ICLR}.

\bibitem[{Rush et~al.(2015)Rush, Chopra, and Weston}]{rush2015neural}
Alexander~M Rush, Sumit Chopra, and Jason Weston. 2015.
\newblock A neural attention model for abstractive sentence summarization.
\newblock In \emph{Proc. EMNLP}.

\bibitem[{Sennrich et~al.(2016)Sennrich, Haddow, and
  Birch}]{sennrich2016neural}
Rico Sennrich, Barry Haddow, and Alexandra Birch. 2016.
\newblock Neural machine translation of rare words with subword units.
\newblock In \emph{Proc. ACL}.

\bibitem[{Vaswani et~al.(2017)Vaswani, Shazeer, Parmar, Uszkoreit, Jones,
  Gomez, Kaiser, and Polosukhin}]{vaswani2017attention}
Ashish Vaswani, Noam Shazeer, Niki Parmar, Jakob Uszkoreit, Llion Jones,
  Aidan~N Gomez, {\L}ukasz Kaiser, and Illia Polosukhin. 2017.
\newblock Attention is all you need.
\newblock In \emph{Proc. NeurIPS}.

\bibitem[{Wan(2011)}]{wan2011using}
Xiaojun Wan. 2011.
\newblock Using bilingual information for cross-language document
  summarization.
\newblock In \emph{Proc. ACL}.

\bibitem[{Wan et~al.(2010)Wan, Li, and Xiao}]{wan2010cross}
Xiaojun Wan, Huiying Li, and Jianguo Xiao. 2010.
\newblock Cross-language document summarization based on machine translation
  quality prediction.
\newblock In \emph{Proc. ACL}.

\bibitem[{Wieting et~al.(2019{\natexlab{a}})Wieting, Berg-Kirkpatrick, Gimpel,
  and Neubig}]{wieting2019beyond}
John Wieting, Taylor Berg-Kirkpatrick, Kevin Gimpel, and Graham Neubig.
  2019{\natexlab{a}}.
\newblock Beyond bleu: Training neural machine translation with semantic
  similarity.
\newblock In \emph{Proc. ACL}.

\bibitem[{Wieting et~al.(2019{\natexlab{b}})Wieting, Gimpel, Neubig, and
  Berg-Kirkpatrick}]{wieting2019simple}
John Wieting, Kevin Gimpel, Graham Neubig, and Taylor Berg-Kirkpatrick.
  2019{\natexlab{b}}.
\newblock Simple and effective paraphrastic similarity from parallel
  translations.
\newblock In \emph{Proc. ACL}.

\bibitem[{Wu et~al.(2016)Wu, Schuster, Chen, Le, Norouzi, Macherey, Krikun,
  Cao, Gao, Macherey et~al.}]{wu2016google}
Yonghui Wu, Mike Schuster, Zhifeng Chen, Quoc~V Le, Mohammad Norouzi, Wolfgang
  Macherey, Maxim Krikun, Yuan Cao, Qin Gao, Klaus Macherey, et~al. 2016.
\newblock Google's neural machine translation system: Bridging the gap between
  human and machine translation.
\newblock \emph{arXiv preprint arXiv:1609.08144}.

\bibitem[{Yao et~al.(2015)Yao, Wan, and Xiao}]{yao2015phrase}
Jin-ge Yao, Xiaojun Wan, and Jianguo Xiao. 2015.
\newblock Phrase-based compressive cross-language summarization.
\newblock In \emph{Proc. EMNLP}.

\bibitem[{Zhang et~al.(2019)Zhang, Kishore, Wu, Weinberger, and
  Artzi}]{zhang2019bertscore}
Tianyi Zhang, Varsha Kishore, Felix Wu, Kilian~Q Weinberger, and Yoav Artzi.
  2019.
\newblock Bertscore: Evaluating text generation with bert.
\newblock \emph{arXiv preprint arXiv:1904.09675}.

\bibitem[{Zhu et~al.(2019)Zhu, Wang, Wang, Zhou, Zhang, Wang1, and Zong}]{ncls}
Junnan Zhu, Qian Wang, Yining Wang, Yu~Zhou, Jiajun Zhang, Shaonan Wang1, and
  Chengqing Zong. 2019.
\newblock {NCLS}: Neural cross-lingual summarization.
\newblock In \emph{Proc. EMNLP}.

\end{thebibliography}
\bibliographystyle{acl_natbib}

\end{document}